# Dense Visual Odometry Using Genetic Algorithm

**Slimane Djema\*[1], Zoubir Abdeslem Benselama[2], Ramdane Hedjar[3], Krabi Abdallah[4]**

**Abstract:** Our work aims to estimate the camera motion mounted on the head of a mobile robot or a moving object from RGB-D images in a static scene. The problem of motion estimation is transformed into a nonlinear least squares function. Methods for solving such problems are iterative. Various classic methods gave an iterative solution by linearizing this function. We can also use the metaheuristic optimization method to solve this problem and improve results. In this paper, a new algorithm is developed for visual odometry using a sequence of RGB-D images. This algorithm is based on a genetic algorithm. The proposed iterative genetic algorithm searches using particles to estimate the optimal motion and then compares it to the traditional methods. To evaluate our method, we use the root mean square error to compare it with the based energy method and another metaheuristic method. We prove the efficiency of our innovative algorithm on a large set of images.

## 1. Introduction

In navigation, odometry is the use of information coming from sensors such as rotary encoders placed in actuators to estimate the position of a moving vehicle. This method has limited use because the wheels can slide on smooth surfaces or spin motionless in the sand, causing an error in the calculation of the movement. This error will accumulate over time and create an increased divergence between the real and estimated positions during the motion.

Visual odometry is the operation of estimating the position using sequential frames coming from an onboard camera analyzed by different algorithms. Visual odometry enhances navigation accuracy in mobile objects using any type of locomotion, such as robots with legs or wheels, on a sticky or sandy surface or drones. This operation improves the performance of the robot for the tasks assigned to it.

In our case, we exploited the dense method, which uses the entire RGB-D image provided by a Microsoft Kinect sensor.

RGB-D cameras refer to digital cameras that provide the colors red, green and blue (RGB) and depth (D) as information for every pixel in the image. The website [1] gives RGBD images and provide associated information as truth trajectory and camera calibration.

In this work, we propose an algorithm for visual odometry that uses a genetic algorithm (GA). This algorithm is presented in Section 3, where we will talk first about modeling the deduction of motion from the vision by a minimization error equation, and then we will explain how to construct the warping image that is used to calculate the error equation. After that, we will talk in detail about the genetic algorithm. In section 4, we present how we compute the relative pose error used to evaluate methods, and in this section, the accuracy of our algorithm is validated with a large set of experiments. The conclusion of our work is given in Section 5.

## 2. Related Work

Visual odometry calculates the object motion using images captured while the robot is moving. Diffirent approaches for visual odometry were distinguished, Among them dense and sparse approaches.

The sparse approach uses feature extraction from the image as noted in [2], [3] or [4] and [5].

A dense approach uses the entire information in the image for motion estimation as in [6]. The first dense approach was presented by [7] which uses image alignment and minimization of geometrical error distance as described in [8] and [9].

Lately, after the discovery of the RGB-D image, the uses of this format have become wide in visual odometry, as given by [10], [11], [12] or [13].

In the literature, there are many optimization methods using metaheuristic approaches explored for the problem of motion estimation by vision using sparse methods such as [14], [15], [16], in addition [17], and [18], [19], or dense methods as in [20] and [21].

In our approach, we estimate the motion by a dense visual odometry method using a metaheuristic algorithm and RGBD images as a data set.

[1]*Department of Electronics, Saad Dahlab University, Blida, Algeria*
[2]*Department of Electronics, Saad Dahlab University, Blida, Algeria*
[3]*Department of Computer Engineering, King Saud University, Riyadh, Saudi Arabia*
[4]*Department of Electronics, Saad Dahlab University, Blida, Algeria*
*\* Corresponding Author Email: djema20132@gmail.com*

## 3. Method

In this part, we explain in a simplified and integrated way the framework for motion estimation between two RGB-D images in a static scene as described in [8] using a genetic algorithm.

### 3.1. System modeling

Visual odometry aims to calculate the motion estimated between two successive frames ($I_t$, $I_{t+1}$) captured by a camera mounted on the head of a mobile robot or a moving object. We will show how this problem has been modeled as a mathematical equation.

An image $r$ called residue is calculated by subtracting the pixel intensity photometric, i.e., the pixel color code average, of each two pixels at the same position from these two images. The pixels in $r$ are more enlightened, engendered a higher error. The following function, as described in [10], represents the intensity error between two consecutive RGB frames of $N$ pixels:

$$E(\xi) = \frac{1}{N} \sum_{i=1}^{N} |I_{t+1}(\omega(\xi, p_i)) - I_t|^2 = \frac{1}{N} \sum_{i=1}^{N} |r_i(\xi)|^2 \quad (1)$$

Where $\xi \in \mathbb{R}^6$ is the motion that we seek to find, and $\omega(\xi, p_i)$ is the warping function.

The motion estimation of the camera consists of minimizing the error of the intensities (also called photometric error) of all pixels of the image. Theoretically, if the motion vector $\xi$ of the camera between the two images is perfectly known, the error of the intensities on all the pixels is null; but in reality, this error is never null because of the noise of the sensor and the changes in the visibility angle of objects, etc. However, this error remains minimal knowing the true motion vector between the two images. The objective of these approaches is to find the camera motion estimated $\xi$ between two images that minimizes the intensities of all the pixels of the image residual by minimizing the following function:

$$\xi = \min_\xi E(\xi) = \min_\xi \frac{1}{N} \sum_{i=1}^{N} |r_i(\xi)|^2. \quad (2)$$

This equation is solved by a genetic algorithm in this work. The warp function is considered an essential part of creating the residual image $r_i(\xi)$.

### 3.2. Building the Warp function

The warp function $\omega(\xi, p)$ in equation (1) changes the position of pixels in $I_{t+1}$ to build a warped image, which we'll capture by a camera if we move on the inverse of the motion $\xi$. Then we subtract this new image from $I_t$ and judge if the motion proposed $\xi$ is optimal by calculating the error equation (1).

The warp function is composed of a set of transformations shown in Figure 1 as noted in [10], [22] and [23]. A pixel $p$ of coordinates $(u; v; d)$ of the image $I_{t+1}$ is projected at a point $M$ in $3D$ of coordinates $(X; Y; Z)$ or $P_M$ by the transformation $P^{-1}$. Then M is transformed from the landmark attached to $I_{t+1}$ to a point M in $3D$ of coordinates $(X'; Y'; Z')$ or $P_{M'}$ in the landmark of $I_{t+1}(\omega(\xi, p_i))$ by the transformation $g(\xi)$ as shown in the following equation:

$$P_{M'} = g(\xi) \times P_M. \quad (3)$$

Finally, $M'$ is projected in the image plane of the warped image $I_{t+1}(\omega(\xi, p_i))$ by the transformation $P$. Thus, the function of warp is written in the following form:

$$\omega(\xi, p) = P(g(\xi)P^{-1}(p)). \quad (4)$$

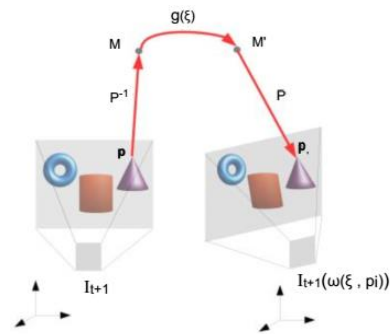

**Fig 1.** The warp function consists of a set of transformations that project each pixel in the image $I_{t+1}$ into warped image $I_{t+1}(\omega(\xi, p_i))$.

Let $P$ be the transformation that makes it possible to go from a $3D$ point to a pixel in the image. Each $3D$ point of coordinates $(X; Y; Z)$ in space is bound to its corresponding $2D$ pixel $(u; v; d)$ by the following equation:

$$P: \mathbb{R}^3 \to \mathbb{R}^2; \quad (X; Y; Z) \to (u, v)$$

$$\begin{cases} u = \frac{X \times f_x}{Z} + c_x. \\ v = \frac{Y \times f_y}{Z} + c_y. \end{cases} \quad (5)$$

It is possible to rebuild the $3D$ point of the scene projected to a pixel thanks to the RGB image type.

The projection $P$ makes it possible to go from a $3D$ scene to a $2D$ frame. Each $3D$ point of coordinates $(X; Y; Z)$ in space is bound to its corresponding $2D$ pixel $(u; v; d)$ by the following equation:

$$P^{-1}: \mathbb{R}^2 \to \mathbb{R}^3; \quad (u, v, d) \to (X; Y; Z)$$

$$\begin{cases} X = \frac{u - c_x}{f_x} \times d \\ Y = \frac{v - c_y}{f_y} \times d \\ Z = d. \end{cases} \quad (6)$$

With *fx, fy, cx* and *cy* are respectively the focal lengths and the optical centers of the camera. *d* is the depth of the pixel returned by the camera.

The motion of a rigid object in *3D* space is perfectly described by its position relative to a fixed reference at all times. *R* and *T* can be grouped into a single rigid transformation matrix *g* as:

$$g = \begin{bmatrix} R & T \\ 0 & 1 \end{bmatrix} \in \mathbb{R}^{4\times4}.$$
(7)

Where $R = \begin{bmatrix} r_{11} & r_{12} & r_{13} \\ r_{21} & r_{22} & r_{23} \\ r_{31} & r_{32} & r_{33} \end{bmatrix}$ is the rotation motion, and $T = \begin{bmatrix} t_x \\ t_y \\ t_z \end{bmatrix}$ is the translation motion.

There is a useful minimalistic representation when, for example, one seeks to determine the parameters using a numerical optimization method. Indeed, each transformation matrix describing the motion of a rigid object has a representation by the vector ξ using six degrees of freedom:

$$\xi = [\, v_1 \ v_3 \ v_2 \ w_1 \ w_2 \ w_3].$$

With $v = (v_1; v_2; v_3)$ the linear velocity and $w = (w_1; w_2; w_3)$ angular velocity.

We use the Lie algebra given in [24] to represent the matrix *g* in the function of the twist coordinates ξ; the transformation *g* can be calculated from ξ using the exponential mapping from Lie Algebra to Lie Group as described in [10] and [25]:

$$g(\xi) = e^{\hat{\xi}}$$
(8)

With $\hat{\xi}$ is the anti-symmetric matrix equal to:

$$\hat{\xi} = \begin{bmatrix} 0 & -w_3 & w_2 & v_1 \\ w_3 & 0 & -w_1 & v_2 \\ -w_2 & w_1 & 0 & v_3 \\ 0 & 0 & 0 & 0 \end{bmatrix}$$

Thus, we talked about all the components of equation (4) as $\xi$, $g(\xi)$, *P*, and $P^{-1}$, which build the warping of each pixel in the image $I_{t+1}$, and so we get $I_{t+1}(\omega(\xi, p_i))$, then we can calculate the residual image and photometric error using the equation (1). As for ξ, several values are proposed by GA as a motion for the particles, after that we calculate the corresponding error value for each particle, as we will explain in our proposed algorithm.

### 3.3. Genetic Algorithm for motion estimation

Genetic algorithms, a type of optimization algorithm, were developed in the 1970s to comprehend the reproduction field of living organisms and the behavior of their genes. Then, they have been applied as an algorithm in machine learning. The main role of the genetic algorithm is to generate several motions based on previously existing motions using their specified equations, and then we will choose the motions that result in less error using the function (1), repeat the same work in the next iterations until the stopping conditions are reached, and finally determine the best motion ever ξ using the function (2).

The explanation of the genetic algorithm (GA) design as mentioned in [26] is the following:

#### 3.3.1. Representation
The position ξ has been identified as a chromosome, and the decision variables in position ξ are genes with 6 alleles; the first three are reserved for the linear velocity and the last three are reserved for the angular velocity.

#### 3.3.2. Population initialization
Each particle of the population must have an initial position ξ using an array of continuous uniform random numbers, each variable is limited by a lower and upper bound.

#### 3.3.3. Objective Function
This function is a mathematical equation, and its maximum value corresponds to the best position estimated in the scene. The term fitness refers to equation (10).

#### 3.3.4. Selection strategy
At this stage, particles are selected for reproduction, and several methods can be used for this. In our experience, we use the roulette wheel selection method, which will give each particle $p_i$ of the population a value of probability $prob_i$ that is proportional to its fitness value, as mentioned in equation (10). We note that $Ei$ is the error of individual $p_i$ and $E_{min}$ is the minimum error, and in the following, the fitness $f_i$ of the particle $p_i$ is

$$f_i = exp(-8 * E_i / E_{min}).$$
(8)

Its probability of being selected is

$$prob_i = f_i / (\textstyle\sum_{i=1}^{n} f_i).$$
(9)

Then we apply the cumulative sum of elements to each selection probability

$$probc_i = cumsum(prob_i).$$
(10)

The random choice of μ particles for mating is made through an indiscriminate variable as an independent spin of the roulette wheel. Better individuals, or the individuals that have the minimum error or the maximum fitness, have more chances to be chosen for the next stage thanks to equation (10).

#### 3.3.5. Reproduction strategy
In our implementation code in Matlab, this strategy consists of two processes to create new particles.

• *Mutation:* this process is done on each particle separately. Thirty percent of the population are selected at random to

undergo the mutation. The mutation rate ($p_m$=0.1) is the probability of a change in particle motion selected for mutation.

The formula for mutation is:

$$\xi'=\xi+M.$$

(11)

and M is a random vector of mutation computed from:

$$M=\delta i*randn(size\_\xi)$$

(12)

where $\delta i$ is calculated by the formula:

$$\delta i = p_m *(\xi_i^U - \xi_i^L )$$

(13)

where $\xi_i^U$(resp.$\xi_i^L$) represents the upper bound (resp. lower bound) for $\xi i$.

• *Crossover:* the resulting parents using the roulette wheel selection method will be used in recombination (The crossover process). The goal of recombination is to create offspring that carry genes from both parents.

Among the crossover processes most commonly used are the intermediate crossover processes. The intermediate crossover attempts to average the positions corresponding to the two parents. The equations of crossover create two individuals $O_1$ and $O_2$ using the weighted average:

$$\begin{cases} O1i = \alpha\xi 1i + (1-\alpha)\xi 2i \\ O2i = \alpha\xi 2i + (1-\alpha)\xi 1i \end{cases}$$

(14)

The extra range factor for crossover α represents the proportion of parent choice as random arrays from the continuous uniform distribution. Then, the new and old particles will create the future population.

### 3.3.6. Replacement strategy

The new offspring and the old particles compete for existence in the future population. We do this by creating a merged population made up of the previous elements and the offspring resulting from crossover and mutation strategies, then we sort order the population using errors calculated by the function (1) and choosing the ones that have the minimum errors according to the required number of particles. Then updating the minimum error ever found.

### 3.4. Pyramid Multi-resolution

Equation (2) is solved by minimizing the intensities of all the pixels of the image residual, and its solution is closer to the truth for tiny motion ξ or small image resolution. To improve the final motion estimation, we present a pyramid as mentioned in [22], [27] and [28], where the down-sampled resolution (DSR) of each image is performed by a factor of 2 (see Figure 2). In the first, we calculate ξ with the image corresponding to high-level DSR (level 4) and this motion will be used as initialization for the next low level in the pyramid up to the initial resolution of the image, where we deduce the optimal motion ξ.

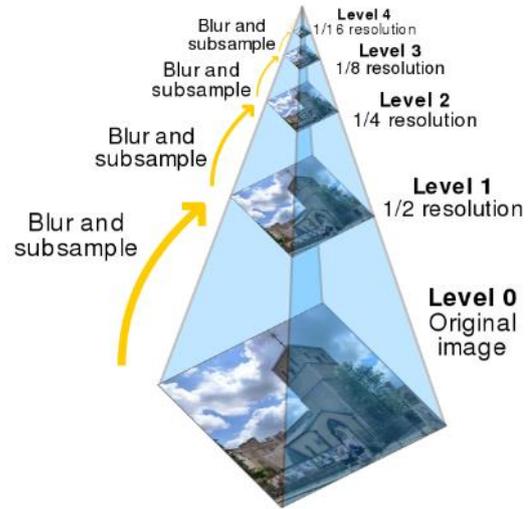

**Fig 2**. Visual representation of an image pyramid with 5 levels.

### 3.5. Stopping criteria

There are a lot of stopping criteria during the execution of code metaheuristics. We used two procedures for stopping below:

• *Static procedure:* the stopping of the execution must be known a priori using a maximum value of iterations for each DSR level of image resolution.

• *Diversity procedure:* the end of the execution of code must be when the best particle error stagnates within a specified number of iterations, keeping the execution of the algorithm useless.

### 3.6. Overall algorithm

In the following, we present the most important step that GA goes through to reach the best motion.

*1. Initialization*

   *a. Set the number of particles as N.*

   *b. Set the number of GA iterations as M.*

   *c. Set the variables bound*

   *d. Set crossover and mutation percentage*

   *e. For i=1,...,N, set $\xi_0^i$=rand(1).*

   *f. Set initial parameters camera intrinsic*

 *2. Main Loop*

*for j=1:M iterations.*

   *-Select parents using the roulette wheel selection*

    *for i=1:ns particles selected (also called Parents)*

    *-Update the particles via Crossover and Mutation*
     *end*

-*Evaluate f(P$_j^i$) and update P$^i$ and get* ξ *of P$_{best}$*
   *end*

*Retenu*ξ

The main role of this algorithm is to generate several motions and calculate the errors produced by equation (1), then deduce the best motion that corresponds to the minimum error. After that, it repeats this operation several iterations until it reaches an optimal motion.

First, this algorithm gives initial motion *ξ* for each item in the group of particles, and then a specific number of particles are selected and used to generate new particles. After that, these new particles replace the old ones, which corresponds to a greater value of error *E(ξ)*. Thus, it has completed one iteration. Then it selects a new group of particles for the next iteration, and follows the same previous steps. After the stopping criteria are met, the motion of the best particle is determined.

We will evaluate this algorithm with several experiments.

## 4. Evaluation
In this section, we evaluate our method for motion estimation on a static scene using RGB-D frames that are available on [1]. For this, we use RPE, RMSE and 3D trajectory to compare our method to particle swarm optimization (PSO), and energy-based as a classic method, which are mentioned respectively in [20] and [12].

### 4.1. Dataset

We get the RGB-D datasets existing on the website [1] from the Kinect camera.

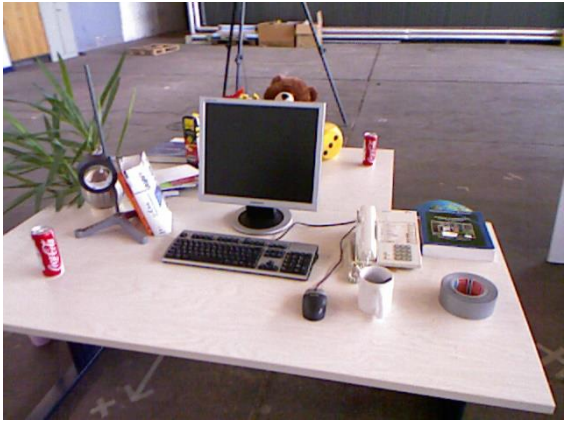

(a)

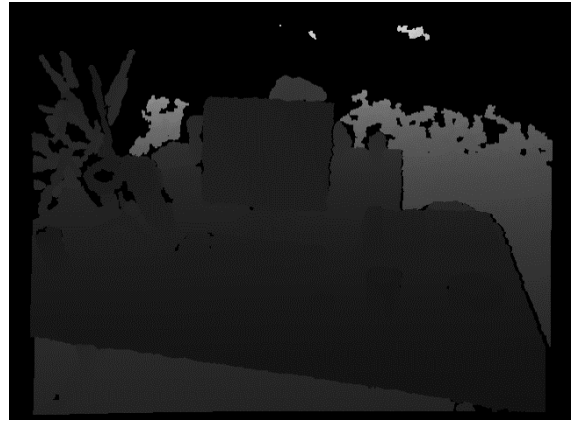

(b)

**Fig 3.** RGB (a) and depth (b) image from "fr2/desk" sequence.

Figure 3(a) represents an RGB frame and Figure 3(b) the corresponding depth, using these two images we have data of 3D scene that formulate an RGB-D image.

### 4.2. Relative pose error

Relative Pose Error (RPE) calculate the drift of the trajectory estimated to the truth trajectory as described in [29], [30] and [31] in time interval Δ at step *i* as

$$E_i = (Q_i^{-1}Q_{i+\Delta})^{-1}(P_i^{-1}P_{i+\Delta}).$$
(15)

From a sequence of *n* images, *m*=*n*−Δ is individual relative pose errors. We define the root mean squared error (RMSE) as

$$RMSE(E_{1:n},\Delta)=(\frac{1}{m}\sum_{i=1}^{m}||trans(Ei)||^2)^{1/2}.$$
(16)

Where *trans(Ei)* is the translational component of RPE.

### 4.3. Real-time graphical user interfaces

Multicriteria optimization is the selection of the best element from some set of available alternatives. The initial parameters of a metaheuristic method like particles number, probability value, and others are considered criteria and we must manipulate them to achieve optimal results, this is what prompted us to build a graphical user interfaces (GUI) in MATLAB shown in Figure4 for various aspects of execution code result in real time to follow the results and choose the best value of criteria to get an optimal solution.

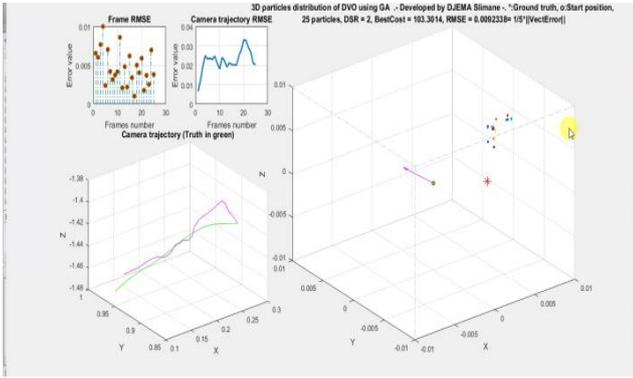

**Fig 4.** The real-time graphical user interfaces view code execution in MATLAB

The graphical user interfaces represent in figure 4 contains four windows, that's On the right there is a three-dimensional space in the form of a parallelogram with dimensions representing the field of motion of the particles, showing us how the particles behave in their search for the value of motion by searching for the lowest error value in equation (1), and in the center of the finite space we notice a green disk representing the three-dimensional position In which the previous image was taken, and the red star represents the location where the next image was taken, and it is the place where the colored particles are looking for, as their color reflects the error value, as whenever the lowest error value or BestCost is in blue, and this allows us to observe the behavior of particles and evaluating the proposed method. As for the red arrow, it represents the direction of the error overlapping with the sequence of images with a three-dimensional ray, and its length is mentioned above. In addition, the number of particles is mentioned and the Down Sampled Resolution DSR and RMSE correspond to the best particle, and this image clearly shows the behavior of the particles in finding the motion between two consecutive photos. The final RMSE between two consecutive images corresponding to the best particle is represented in the form of impulses as shown in the upper graphical example, and on the right is a graph corresponding to camera trajectory RMSE, either at the bottom, a graph represents the truth and estimated trajectory in the same 3D scene.

### 4.4. Experimental setup

We evaluated our algorithm through various experiments in a static environment using RGB-D images of dimension 640×480 with a frame rate of 30Hz. These images and their corresponding ground truths are available on the website [1].

In our first experiment, we used 90 consecutive frames of *rgbd_dataset_freiburg1_xyz*. The function (17) gives the distance error between estimated motion and ground truth. Therefore, using this function, we compute the camera trajectory error of different methods through 90 consecutive images, and we represent these results in the same graph in figure 5. The evolution of distance error indicates that the accumulative error of 90 frames related to the classic method is the least, but the quasi-stabilization of the error of the PSO method in the last 60 frames allowed it to outperform RMSE as we can see in table 1, which is considered the most important evaluation criterion in visual odometry. Although the final accumulation error of GA is greater, the error almost preserved its value between the beginning and the end of the last thirty frames, which helped him improve the value of translational components of RPE and thus outperform the RMSE of classic method.

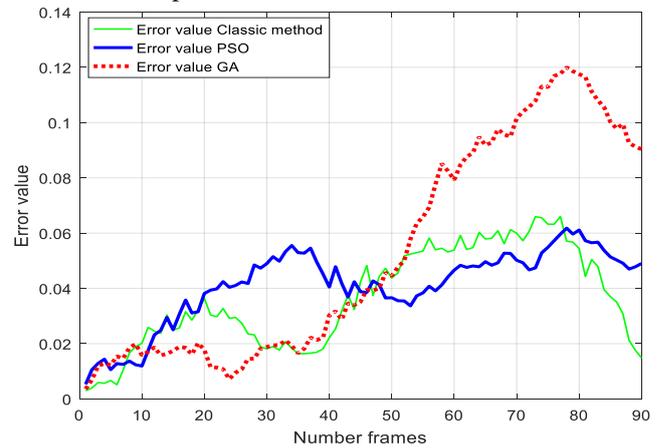

**Fig 5.** Camera trajectory error of GA, PSO and classic method using a part of fr1_xyz dataset.

The representation in the 3D scene of the truth and estimated camera trajectory clearly shows the effectiveness of the motion estimation methods. Figures 6, 7 and 8 show the camera trajectory for fr1_xyz of ground truth and different motion estimation methods; classic and metaheuristics methods.

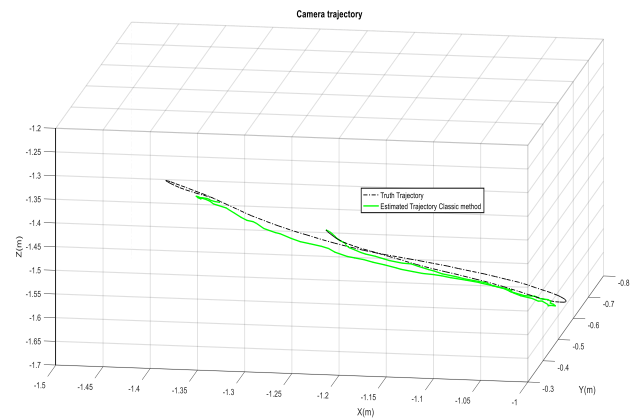

**Fig 6.** Truth camera trajectory and classic method using a part of fr1_xyz dataset.

Through ninety RGBD frames and in a back and forth path, the three methods gave very acceptable results, and this confirms that the error of a representation in Figure 5 is very small compared to the path traveled, and there was no deviation away from the true trajectory.

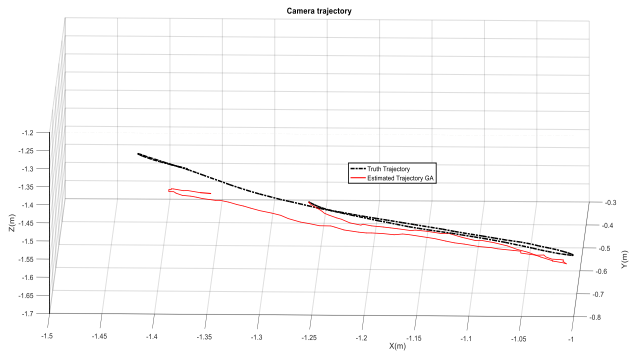

**Fig 7.** Truth camera trajectory with GA method using a part of fr1_xyz dataset.

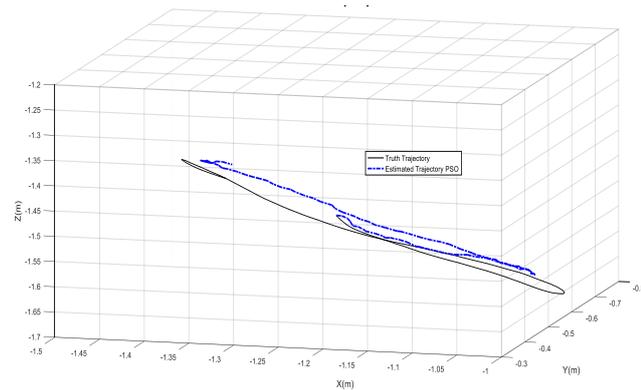

**Fig 8.** Truth camera trajectory with PSO method using a part of fr1_xyz dataset.

In the second experiment, we used 60 consecutive frames of *rgbd_dataset_freiburg2_desk*. The distance error between the ground truth and the trajectory of GA, PSO and the classic method is represented in Figure 9. We notice that in the first thirty frames, the GA method achieved the least distance error compared to the other methods, and the results were close compared to the classic method. This is clearly shown in the Table I with the superiority of the GA.

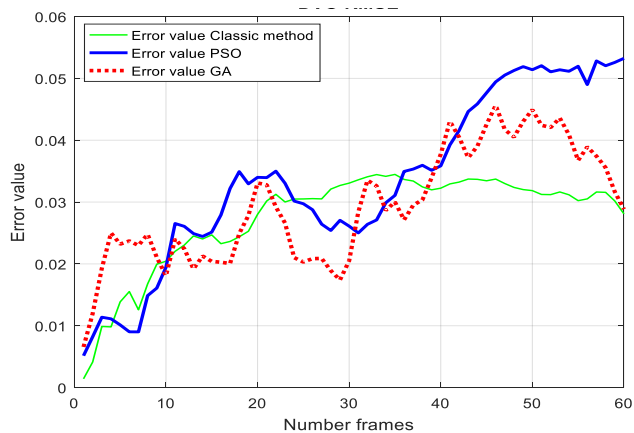

**Fig 9.** Camera trajectory error of GA, PSO and Classic method using a part of fr2_desk dataset.

The representation in the same 3D scene of the truth and the estimated camera trajectory clearly shows the effectiveness of the motion estimation method. Figure 10 shows the camera trajectory for fr2_desk of ground truth and different motion estimation methods; classic and meta-heuristics methods. The three methods gave acceptable results, where we notice the corresponding trajectories are very close to the true trajectory, but for the trajectory corresponding to GA, we notice that it is the closest to the truth trajectory, which confirms through this experiment that this innovative method competes with the previous methods; classic and PSO or even better, and this confirms previously obtained results from Figure 10 and Table 1.

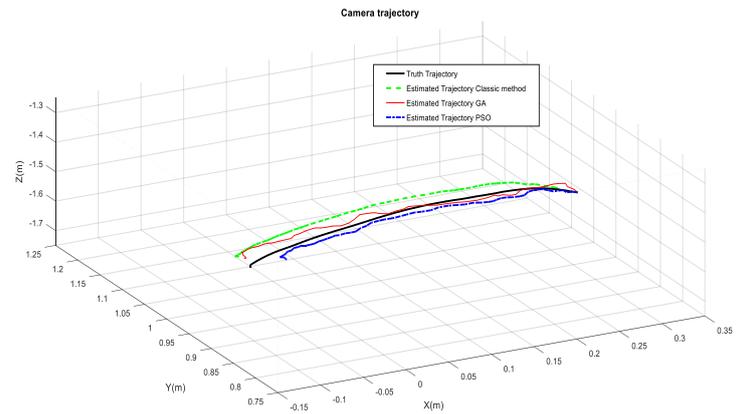

**Fig 10.** Truth camera trajectory with GA, PSO and Classic method using a part of fr2_desk dataset.

Table 1 shows the root mean square error (RMSE) calculated using the function (18) for two previous methods GA, PSO and the results of the methods classic tested in [10].

**Table 1.** Root mean square error (RMSE) of drift in meters per second for different methods for ground truth.

| Dataset | GA | PSO | Classic |
|---------|------|------|---------|
| *fr1_xyz* | 0.04062 m | 0.03598 m | 0.04827 m |
| *fr2_desk* | 0.01856 m | 0.02836 m | 0.02524 m |

For the dataset (freiburg1_xyz) our method using GA has proven its efficacy in comparison to the classic method. Regarding the dataset (freiburg2_desk), our innovative method using GA has successfully proven its efficiency compared to both methods used in this experiment: PSO and the classic method.

## 5. Conclusion and Future Work

We built a new algorithm to estimate the trajectory of a camera in a static scene using a metaheuristic method. This method is a genetic algorithm. After a large set of experiments, we have demonstrated the efficacy of this method and the improvement in results. Gradually, we showed how to achieve the desired results. In the future, we want to extend this work to estimate body motion in a dynamic scene.